\documentclass[letterpaper,10pt,conference]{ieeeconf}
\IEEEoverridecommandlockouts
\usepackage[T1]{fontenc}
\usepackage[utf8]{inputenc}
\usepackage{microtype}
\usepackage{amsmath,amssymb}
\usepackage{graphicx}
\usepackage{booktabs}
\usepackage{xcolor}
\usepackage{url}
\usepackage[hidelinks]{hyperref}
\usepackage{comment}
\graphicspath{{figures/}}
\newenvironment{IEEEkeywords}{\begin{keywords}}{\end{keywords}}
\newcommand{\SRret}{\mathrm{SR}_{\mathrm{ret}\mid\mathrm{out}}}
\newcommand{\ARret}{\mathrm{AR}_{\mathrm{ret}\mid\mathrm{out}}}

\begin{document}
\title{\LARGE \bf Reliability-Aware Sparse Route Memory for\\
Round-Trip Vision-Language Navigation}

\author{Bojun Long, Lingfan Bao, Tianhu Peng, Jingcheng Sun, and Chengxu Zhou%
\thanks{Bojun Long, Lingfan Bao, Tianhu Peng, Jingcheng Sun, and Chengxu Zhou
are with the Department of Computer Science, University College London,
London, United Kingdom.
Emails: \{bojun.long.25, lingfan.bao.24, tianhu.peng.24,
jingcheng.sun.16, chengxu.zhou\}@ucl.ac.uk.}%
\thanks{This work was partially supported by the Advanced Research and Invention Agency (grant number SMRB-SE01-P06) and the NVIDIA Academic Grant Program.}%
}
\maketitle
\thispagestyle{empty}
\pagestyle{empty}

\begin{abstract}
Vision-language navigation (VLN) is typically evaluated as a one-way task, although deployed robots may need to return after reaching a goal. We study continuous round-trip VLN and diagnose failures in directional observability, deviation recovery, and termination stability. We propose a reliability-aware sparse route memory that records the executed Outbound trajectory as ordered geometric anchors and queries them in reverse through a structured hint, action-level arbitration, and terminal verification. On 50 reverse-paired episodes using NaVILA and a simulated Unitree Go2, language-only Return succeeds in 22.0\% of episodes, while our online system reaches 55.1\%. With exact route information, the same interfaces achieve 86.0\%, showing that effective Return requires both accurate information and consistent action on that information. The remaining online gap arises mainly from geometric evidence that is too unreliable to authorise intervention. These results distinguish information quality, behavioural consistency, and online reliability as separate limits in long-horizon navigation.
\end{abstract}

\section{Introduction}
Vision-language navigation (VLN) requires an embodied agent to interpret a
natural-language instruction and navigate to its target. Existing benchmarks
predominantly end when a one-way goal is reached. Practical mobile robots,
however, may need to return to an operator or charging point after completing
an inspection or retrieval task. On the same physical routes, a
standalone one-way traversal succeeds 40\% of
the time, yet the identical route executed as the Return leg of a continuous
round trip succeeds in only 22.0\%. Return is therefore not
a categorically different task but the same route made harder by continuous
execution: its initial state is produced by Outbound execution, the longer
horizon lets errors accumulate, the route is visually asymmetric in reverse, and
success depends on recognising the original start and terminating there.

The same setting also supplies an opportunity. The robot already possesses a
route that it has just traversed. This executed route is available without a
separate mapping or teaching run and contains precisely the spatial prior
needed for Return. The central question is therefore not only how to retain
this prior, but how it should influence a vision-language-action policy during
execution.

Most work improves VLN by supplying stronger representations, memories, or
localisation, implicitly assuming that sufficiently accurate information will
produce correct behaviour. We test this assumption directly. A controlled
Oracle condition removes localisation error while retaining the same route
memory and interfaces as the online system. It thereby separates two otherwise
confounded factors: the quality of the external reference and the consistency
with which the policy acts on it. The subsequent online experiment then
measures the reliability gap to this information-quality ceiling.

We organise Return failure into three execution-level dimensions:
\emph{directional observability} (D1), the absence of a persistent Return
reference; \emph{recovery from deviation} (D2), the accumulation of actions
that conflict with a valid direction; and \emph{termination stability} (D3),
the absence of an independent constraint on the arrival judgement. Our route
memory addresses them through a structured route hint, action-level
arbitration, and terminal verification, respectively, while retaining NaVILA
as the primary controller.

This paper makes three contributions. First, it introduces a continuous Outbound--Return VLN protocol
using reverse-paired human instructions and quantifies the performance
degradation caused by continuous execution through a comparison between
standalone one-way traversal and continuous Return on M50. Second, it proposes a sparse geometric memory of the route actually
executed, exposed through three interfaces (hint, action arbitration, terminal
verification) and gated online by a learned, calibrated reliability estimate. Third, by
comparing the Oracle and online settings, we show that exact route information
alone is insufficient for the current policy to produce reliable return
behaviour. We further quantify the gains from action arbitration and termination
verification, and attribute the remaining online performance gap to the limited
availability of reliable geometric evidence. An additional
geometry-only alternative-controller experiment shows that a deterministic route
follower without the VLM performs significantly worse than the full online
system.

\section{Related Work}
\textbf{VLN and extended-horizon execution.}
R2R introduced instruction following over discrete Matterport3D navigation
graphs~\cite{anderson2018r2r}, and VLN-CE extended the task to continuous
environments with collision and local control~\cite{krantz2020vlnce}. NaVILA
couples a vision-language-action model to a legged locomotion policy for
real-time execution~\cite{cheng2025navila}. Long-horizon VLN further exposes
memory and progress-estimation demands~\cite{song2025longhorizon}, while
CorrectNav and BudVLN improve recovery from policy-generated deviations through
training~\cite{yu2025correctnav,he2026budvln}. In contrast, we study a
continuous round trip and intervene at execution time without modifying the
base policy.

\textbf{Spatial memory and route reuse.}
Explicit memories such as DUET, ETPNav, and Mem2Ego preserve topological or
retrievable spatial context outside a model's hidden state
~\cite{chen2022duet,an2024etpnav,zhang2025mem2ego}. Visual Teach-and-Repeat
instead records a route during teaching and localises against it during later
traversals~\cite{furgale2010vtr}. Our memory is task-specific rather than a
general environment map: it is constructed online as a by-product of a
language-directed Outbound journey and queried immediately in reverse, while
semantic decisions remain with the VLN policy.

\textbf{Relocalisation, reliability, and runtime intervention.}
Single-frame place recognition is vulnerable to perceptual aliasing; sequence
matching and temporal filtering improve route identity under repeated
appearance~\cite{milford2012seqslam,tomita2022sequence}. Local point-cloud
registration can likewise appear accurate in geometrically degenerate
environments, motivating observability-aware registration
~\cite{tuna2022xicp,tuna2024degeneracy}. Separately, VLN termination is a known
source of failure~\cite{zhao2023mindgap,xiang2020learningtostop}. Runtime
filters and residual policies constrain learned actions while preserving the
base controller in nominal operation
~\cite{wabersich2021predictive,johannink2019residual,lin2023text2motion}. We
connect these directions by using reliability estimates not merely to select a
route location, but to authorise distinct linguistic, action, and terminal
interfaces at each decision step.

\section{Problem Formulation and Method}
\label{sec:method}
\subsection{Continuous Round-Trip VLN}
Let a one-way episode be $E_i=(S_i,P_i,x_i)$, denoting its scene, reference
path, and human instruction. We pair episodes from the same scene whose paths
approximately reverse one another:
\begin{equation}
E^{\mathrm{RT}}=(E^{\mathrm{out}},E^{\mathrm{ret}}),\qquad
P^{\mathrm{ret}}\approx\operatorname{reverse}(P^{\mathrm{out}}).
\end{equation}
The robot follows $x^{\mathrm{out}}$, performs a deterministic $360^\circ$
scan after a successful Outbound phase, and then follows
$x^{\mathrm{ret}}$ back towards the original start $s_0$. There is no
teleportation or environment reset: Return begins from the actual terminal
state of Outbound execution. Using an existing reverse instruction avoids the
confound introduced by automatically generated language.

For $N$ episodes, let $y_e^{\mathrm{out}}$, $y_e^{\mathrm{ret}}$, and
$y_e^{\mathrm{arr}}$ indicate Outbound success, Return success, and whether
the final Return position lies within the $\tau=3.0$~m success radius. We use
\begin{align}
\SRret &=\frac{\sum_e y_e^{\mathrm{out}}y_e^{\mathrm{ret}}}
                  {\sum_e y_e^{\mathrm{out}}}, \\
\ARret &=\frac{\sum_e y_e^{\mathrm{out}}y_e^{\mathrm{arr}}}
                  {\sum_e y_e^{\mathrm{out}}}, \\
\Delta_{\mathrm{term}}&=\ARret-\SRret .
\end{align}
Success requires both arrival and a valid STOP. The termination deficit
$\Delta_{\mathrm{term}}$ therefore counts episodes that finish inside the
start region without terminating successfully and separates D3 from failures
of the Return trajectory.

\subsection{Sparse Route Memory}
During Outbound, NaVILA follows the original instruction while a background
module creates an anchor whenever travelled path distance exceeds $\delta_a$.
The resulting memory
\begin{equation}
\mathcal{M}=\{A_0,\ldots,A_K\},\qquad
A_i=(G_i,i,\ell_i,T_{i-1,i})
\end{equation}
stores local 2-D LiDAR geometry $G_i$, topological index $i$, cumulative route
distance $\ell_i$, and the relative transformation $T_{i-1,i}$ between
neighbouring anchors. It represents only the executed trajectory rather than a
global map. Once Outbound succeeds, the memory is frozen and queried in the
order $A_K,\ldots,A_0$.

\begin{figure*}[t]
    \centering
    \includegraphics[width=0.96\textwidth]{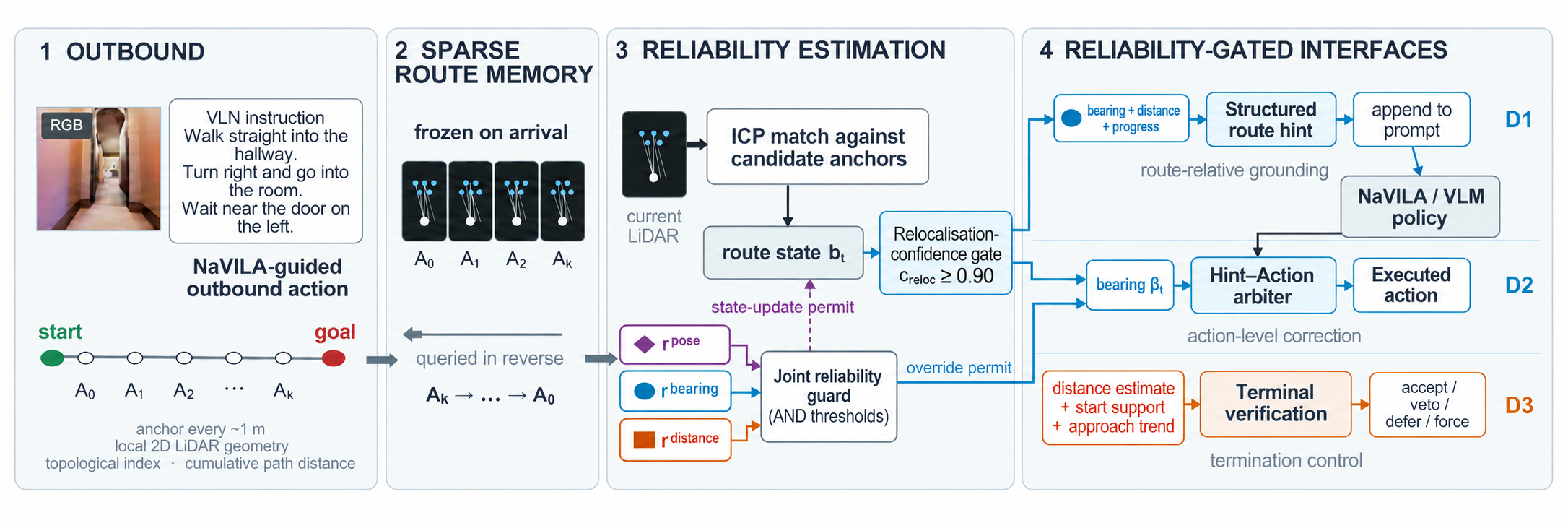}
    \caption{Reliability-aware sparse route memory. Outbound execution records
    ordered geometric anchors. During Return, online registration
    estimates a route state, a relocalisation-confidence gate for the structured
    hint (D1) and Hint-Action arbiter (D2), and a learned reliability guard on
    route-state updates and overrides. NaVILA remains the primary controller.}
    \label{fig:system}
\end{figure*}

\subsection{Reliability-Aware Return Navigation}
At Return step $t$, the system selects candidate anchors near the previous
route state and registers the current LiDAR geometry against them. Normal
tracking considers the current and next reverse anchor; failure expands this
to a bounded recovery set. Because repeated corridors can register well to an
incorrect location, the route state is updated only when anchor topology,
multi-frame consistency, and candidate ambiguity provide sufficient support.
Otherwise the previous state is retained rather than jumping to the strongest
single-frame match.

Reliability is decomposed by the geometric quantity whose error matters:
\begin{equation}
r_t=(r_t^{\mathrm{pose}},r_t^{\mathrm{bearing}},r_t^{\mathrm{distance}}).
\end{equation}
Each component is a calibrated probability that the corresponding
signal lies within tolerance. A gradient-boosted model maps registration
features---raw ICP fitness, inlier count, overlap and residuals, multi-basin
scores and margins, Scan-Context yaw and region measures, localizability
eigenvalues, and temporal windows over 4--32 frames---to a raw score that Platt
scaling calibrates to \mbox{$[0,1]$}; the three heads share this feature vector but
have separate models. In the deployed system the components are combined
conservatively: route-state updates and action overrides are authorised only when all three reliability components are jointly trusted, using thresholds \mbox{$r^{\mathrm{bearing}}\!\ge\!0.68$},
\mbox{$r^{\mathrm{distance}}\!\ge\!0.74$}, and \mbox{$r^{\mathrm{pose}}\!\ge\!0.70$}
(equivalently, calibrated bad-probabilities below \mbox{$0.32/0.26/0.30$}). This joint
gate acts as a route-state guard on updates and overrides; the hint and arbiter
are additionally gated by a coarser relocalisation-confidence threshold
(Sec.~\mbox{\ref{sec:setup}}). Per-component specialisation to individual operations
is a design target the current implementation does not yet separate.

\textbf{Structured route hint (D1).}
When $r_t^{\mathrm{bearing}}$ is sufficient, a fixed template appended to the
Return instruction reports the target anchor, its bearing and distance, and
remaining route progress. The bearing is expressed as \texttt{ahead},
\texttt{turn left}, or \texttt{turn right}, with its Cartesian displacement
also included. When evidence is insufficient, the hint is withheld and NaVILA
uses only its RGB observation and the original instruction. This interface
supplies information without constraining the output.

A representative instance is:
\begin{quote}
\footnotesize\ttfamily\raggedright
[System Hint: route anchor A10 is 0.58 m away, ahead; remaining route via
anchor is 10.66 m; vector dx=0.58 m, dy=0.02 m.]
\end{quote}
The same template is used for every hinted configuration, so Oracle and online
runs differ in the source and reliability of the quantities rather than their
linguistic encoding.

\textbf{Hint-Action arbitration (D2).}
Let $a_t^{\mathrm{VLM}}$ be NaVILA's proposed action, $\beta_t$ the reliable
bearing to the target anchor, and $\mathcal{O}_t$ current LiDAR occupancy. The
executed action is
\begin{equation}
a_t=\begin{cases}
a_t^{\beta}, & \substack{a_t^{\mathrm{VLM}}\text{ conflicts with }\beta_t,\\
               \beta_t\text{ is traversable}},\\
a_t^{\mathrm{VLM}}, & \text{otherwise},
\end{cases}
\end{equation}
provided bearing reliability exceeds its threshold. Conflict is evaluated
after quantising both bearing and action into forward, left, and right. An
override is declined if a narrow current-frame occupancy corridor does not
confirm the hinted direction as traversable. Arbitration is therefore
selective: it interrupts clearly inconsistent local actions without replacing
the policy in nominal operation.

Opposite turns and any turn proposed for a forward target are conflicts; a
forward proposal conflicts with a turning target only when
$|\beta_t|\geq30^\circ$, leaving smaller deviations to the policy. Replacement
actions are fixed within each directional class rather than selected as the
discrete action closest to $\beta_t$. Arbitration is disabled within 0.35~m of
the target anchor, where bearing becomes unstable. These choices make the
external signal a conservative correction rather than a local planner.

\textbf{Terminal verification (D3).}
A STOP proposal is treated as a request and checked against distance, start
anchor support, and recent approach trend. The verifier accepts supported
requests, vetoes contradictory ones so navigation continues, defers when
evidence is insufficient, and forces STOP after consistent evidence that the
robot lies inside the start region. Veto and deferral can change where a
trajectory ends, while forced termination converts arrival without STOP into
success.

\section{Experimental Setup}
\label{sec:setup}
Experiments use VLN-CE-Isaac with Matterport3D scenes, a simulated Unitree Go2,
and the 8-bit NaVILA policy on an RTX 4090. Reverse pairs are constructed using
ordered nearest-neighbour matching between reference paths, with a 2.0~m
waypoint tolerance and 0.8 coverage. Their Return endpoints differ from $s_0$
by 1.77~m on average (maximum 5.52~m), an offset shared by every configuration.
Both phases require a valid STOP within 3.0~m. Return success is
measured as the simulator distance from the final position to the Outbound start
\mbox{$s_0$}, not to the reverse episode's goal. Each configuration executes its
own Outbound leg; paired tests consequently use only episodes that succeed
Outbound under both configurations. Outbound success counts differ
across configurations (50, 43, 44, 43, 49) because execution is not
bit-deterministic between runs: although decoding is greedy, the 8-bit policy and
simulator GPU execution introduce run-to-run divergence, so about 19 of the 50
episodes change Outbound outcome across runs while 31 always succeed. All Return
contrasts are therefore computed on per-configuration common-Outbound subsets.

We use two cohorts. A performance-independent pilot pool of 38 episodes checks
the reproduced one-way baseline. All principal results use M50, a cohort of 50
reverse-paired episodes selected from 264 candidates for high historical
Outbound reliability (94.9\%), so the analysis focuses its compute on Return.
This selection makes absolute end-to-end round-trip success optimistic but does
not select on Return outcome; conditional and within-episode paired contrasts
remain the focus.

To obtain a standalone one-way reference on the same M50 routes, we
also execute the 50 reverse-route episodes independently under the
language-only configuration. Each run starts directly from the reverse
episode's initial state and uses its original human instruction, without a
preceding Outbound phase or access to route memory. The NaVILA checkpoint,
action budget, and requirement for a valid STOP within 3.0~m are kept
consistent with the continuous-Return evaluation.

Five configurations share the same policy, episodes, instructions, memory, and
success criterion. \emph{Language-only} disables the memory. \emph{Oracle Hint}
computes anchor bearing and distance from simulator pose and supplies only the
textual hint. \emph{Oracle Hint-Action} additionally enables arbitration, and
\emph{full Oracle} also enables terminal verification. The proposed
\emph{online} system activates the same three interfaces as full Oracle but
estimates bearing and distance by 2-D ICP and applies reliability gating.

We additionally evaluate a \emph{geometry-only route-following}
alternative controller to test whether the geometry of the executed route can
directly replace the semantic policy during Return. NaVILA still executes the
Outbound and confirmation stages. During Return, however, the configuration
removes VLM queries, structured hints, and action arbitration, and instead uses
the accumulated trajectory breadcrumbs, the same sequential-pair ICP route-state
estimate as the online system, and a deterministic reverse-polyline follower.
The follower uses a 0.65~m lookahead, \mbox{$28^\circ/12^\circ$} turn-entry/exit
hysteresis thresholds, and fixed forward and turning commands, while retaining
terminal verification. Stuck recovery is a Return-only fallback used
by the online system: after eight consecutive VLM queries with under 0.15~m of
displacement while the estimated home distance still exceeds 5~m and NaVILA is
not stopping, it turns roughly \mbox{$180^\circ$} in \mbox{$45^\circ$} increments, drives
forward until 0.8~m of net progress, and re-faces the next anchor, for at most
five attempts. It only overrides the action and never writes relocalisation,
reliability, or STOP state. It is enabled only for the online system during
Return; Language-only, all Oracle rows, and this geometry-only controller have
it disabled.
 It is therefore treated as a complete
alternative-controller baseline rather than as a single-factor ablation of the
VLM.

Anchors are spaced by 1.0~m and contain at most 512 points after 0.10~m voxel
downsampling. ICP uses at most 16 iterations, a 0.45~m correspondence threshold,
and multiple angular seeds. The hint and arbiter are authorised when
the relocalisation confidence exceeds 0.90; the learned reliability guard on
route-state updates and overrides uses \mbox{$r^{\mathrm{bearing}}/r^{\mathrm{distance}}/r^{\mathrm{pose}}$}
thresholds of \mbox{$0.68/0.74/0.70$}.
Arbitration uses 15$^\circ$ and 30$^\circ$ forward/conflict thresholds, a
1.0-by-0.84~m traversability corridor, and is disabled within 0.35~m of an
anchor. All parameters were fixed before M50 evaluation.

We report two-sided exact McNemar tests on paired binary outcomes. The main
analysis additionally uses per-step reason codes recorded by each interface.

\section{Results and Diagnostic Analysis}
\subsection{Continuous Return Introduces an Execution Deficit}

On M50, NaVILA succeeds in $20/50=40.0\%$ of the standalone
language-only one-way runs, compared with $11/50=22.0\%$ when the
corresponding routes are executed as the Return phase of a continuous round
trip.

On the pilot pool, NaVILA succeeds Outbound in $17/38=44.7\%$ of episodes,
close to the reported 50.2\% Go2-Vision result despite differences in episode
selection and evaluation details~\cite{cheng2025navila}. Only $5/17=29.4\%$
then succeed Return, consistent with the 22.0\% M50 baseline and indicating
that the reproduction has not suffered a gross loss of one-way capability.

Trajectory logs associate the drop with recurring closed-loop instabilities
rather than one unique failure. Runs oscillate between opposing turns, sustain
directional deviation, stagnate, or terminate incorrectly. These behaviours
also occur occasionally Outbound; continuous Return increases the horizon and
the opportunity for them to be triggered and accumulated.

We first control the most immediate alternatives on one reverse pair. The
reverse neighbour succeeds as an independent one-way episode, stopping 0.317~m
from its goal. The same human instruction and physical route fail as the Return
of its paired episode, even after position and heading are reset exactly to the
expert endpoint; NaVILA stops 11.351~m from $s_0$. Thus neither route
executability, generated language, nor Outbound terminal pose error is a
sufficient explanation.

We then test whether this effect generalises by comparing geometrically matched
routes under two contexts. Of
24 pairs whose round-trip execution enters Return, the same physical route is
completed in $11/24=45.8\%$ of standalone one-way runs but only $3/24=12.5\%$
when embedded as the Return leg. Nine routes change from standalone success to
Return failure and one changes oppositely ($p=0.021$), across nine scenes.
The evidence therefore localises the additional burden to continuous execution
rather than an inherently unexecutable route or automatically generated
language. It does not isolate a single internal cause: phase transition,
accumulated visual context, and the longer chain of decisions remain plausible
contributors.

\subsection{Oracle Analysis: Information Versus Behaviour}
Table~\ref{tab:main_results} presents the central ablation. Supplying exact
bearing and distance solely through the hint raises $\SRret$ from 22.0\% to
37.2\%. On the 43 episodes entering Return in both runs, however, the paired
change from 23.3\% to 37.2\% is not significant ($p=0.180$). Almost two-thirds
of trials therefore still fail when localisation error has been removed.

\begin{table}[t]
\centering
\caption{Conditional Return results on M50. The first four rows form the
Oracle ladder. Paired tests use the common-Outbound subset and compare with the
preceding Oracle row; the online row compares with full Oracle.
The \mbox{$\SRret\,(n{=}31)$} column reports success on the 31 episodes that
succeed Outbound under all five configurations; arrival is not separately
recoverable on this subset.}
\label{tab:main_results}
\footnotesize
\setlength{\tabcolsep}{2.4pt}
\begin{tabular}{@{}lrrrrr@{}}
\toprule
Configuration & Out. & $\ARret$ & $\SRret$ & $\SRret$ & paired $p$ \\
              & ep.  & (\%) & (\%) & $n{=}31$ & \\
\midrule
Language-only          & 50 & 24.0 & 22.0 & 25.8 & --- \\
$+$ Oracle hint        & 43 & 39.5 & 37.2 & 38.7 & 0.180 \\
$+$ action arbitration & 44 & 81.8 & 70.5 & 67.7 & 0.008 \\
$+$ terminal verifier  & 43 & 90.7 & 86.0 & 83.9 & 0.125 \\
\midrule
Reliability-aware online & 49 & 59.2 & 55.1 & 61.3 & 0.027 \\
\bottomrule
\end{tabular}
\end{table}

The hint clearly affects computation: among episodes with common Outbound
success, median trajectory deviation between baseline and Hint runs increases
from 0.10~m during Outbound to 1.08~m during Return ($p=2.4\times10^{-8}$).
Nevertheless, this cannot distinguish correct interpretation followed by
inconsistent execution from a response to an unfamiliar prompt format. Under
either interpretation, accurate information delivered through the linguistic
interface alone is insufficient for the present policy.

\begin{figure*}[t]
    \centering
    \includegraphics[width=0.96\textwidth]{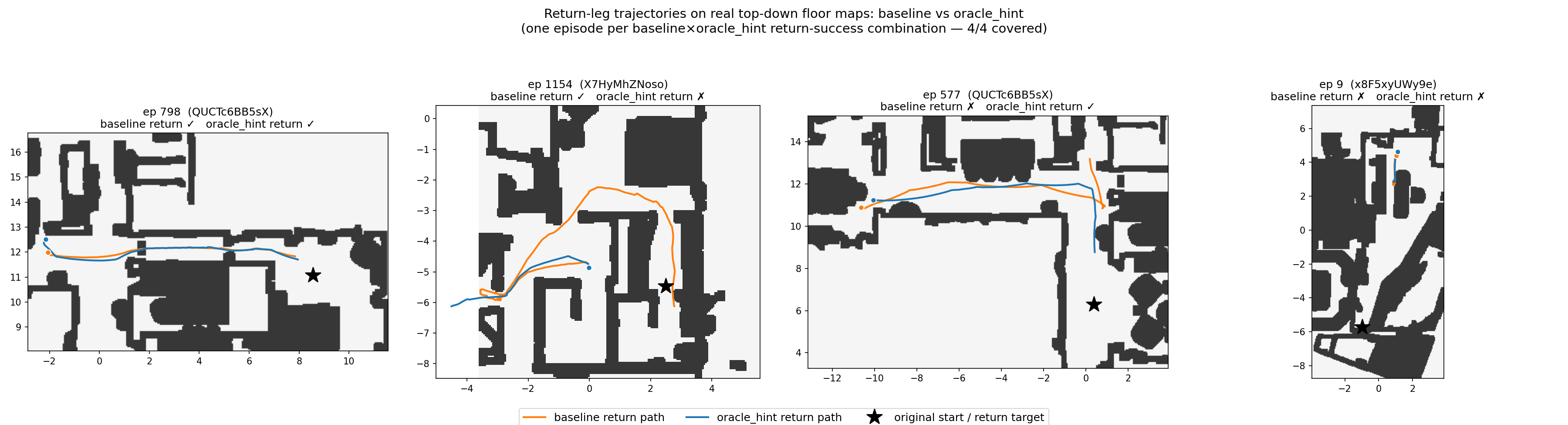}
    \caption{Return trajectories under language-only (orange) and Oracle Hint
    (blue), one episode per combination of the two outcomes. Stars mark $s_0$.
    The Outbound trajectories coincide to numerical precision, so Return starts
    identically and the hint is the remaining experimental difference. The
    rightmost case fails under both configurations, showing that the hint
    alters execution independently of the final outcome.}
    \label{fig:hint_effect}
\end{figure*}

This conclusion is robust to different Outbound endpoints. Restricting the
analysis to the 41 Return legs starting within 1.0~m of one another gives
median Outbound and Return deviations of 0.09 and 0.77~m, respectively
($p=5.3\times10^{-8}$). In 19 episodes the Outbound trajectories coincide to
numerical precision, yet nine Return pairs diverge by more than 0.5~m and by as
much as 6.5~m. Figure~\ref{fig:hint_effect} illustrates four such cases. The
hint changes the executed path even when both runs ultimately fail, but this
behavioural sensitivity is not evidence that its directional semantics were
correctly recovered.

Adding arbitration while holding the exact reference, anchor sequence, and
hint fixed raises success to 70.5\%; the strictly paired Hint-to-Hint-Action
comparison is 32.4\% versus 64.9\% ($p=0.008$). The decision logs explain the
gain. Of 1,604 Return steps, 32.9\% contain an action that conflicts with the
correct anchor direction. Traversability permits 277 overrides (17.3\% of all
steps), while 250 conflicts are declined; hence 82.7\% of model outputs remain
unchanged. The mechanism is not conventional route following, but a selective
constraint that corrects approximately one decision in six.

\begin{table}[t]
\centering
\caption{Arbiter outcomes over Return decisions. Online uses the full-run denominator; Online (authorised) includes only the 492 steps with authorised bearing evidence and the target anchor outside the near-anchor exclusion zone, enabling comparison with Oracle.}

\label{tab:arbiter}
\footnotesize
\setlength{\tabcolsep}{3pt}
\begin{tabular}{@{}lrrr@{}}
\toprule
Outcome & Oracle & Online & Online \\
        & ($n{=}1604$) & ($n{=}1579$) & auth. \mbox{$n{=}492$} \\
\midrule
Action consistent                    & 60.8\% & 14.9\% & 48.0\% \\
Conflict, not traversable            & 15.6\% &  7.2\% & 25.6\% \\
Conflict, overridden                 & 17.3\% &  8.2\% & 26.4\% \\
Target anchor too close              &  6.3\% &  6.8\% & n/a \\
\midrule
Relocalisation confidence withheld   & --- & 62.0\% & --- \\
Reliability guard blocked override   & --- &  0.8\% & --- \\
\bottomrule
\end{tabular}
\end{table}

\begin{figure}[t]
    \centering
    \includegraphics[width=0.96\columnwidth]{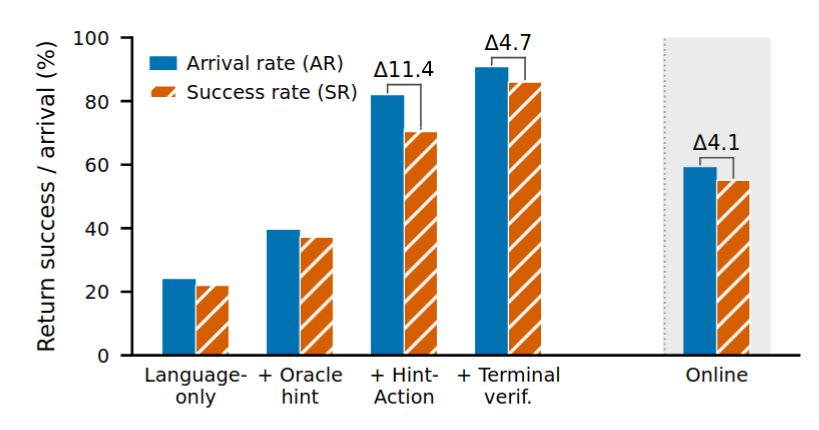}
    \caption{Arrival and Return success conditional on Outbound success. Their
    gap is the termination deficit. Arbitration exposes this deficit, while
    terminal verification both improves arrival and closes most of the gap.}
    \label{fig:arrival}
\end{figure}

Terminal verification raises success further from 70.5\% to 86.0\%. This
15.5-point gain has two components (Fig.~\ref{fig:arrival}). First, arrival
increases from 81.8\% to 90.7\% because vetoed or deferred premature STOP
requests allow navigation to continue. Second, the termination deficit falls
from 11.4 to 4.7 points because forced STOP converts unsupported termination
after arrival into success. Thus D3 is not confined to recognising that the
robot has already arrived: a premature judgement can truncate execution before
arrival. The adjacent paired comparison is not significant at this sample size
($p=0.125$), although the complete Hint-to-full-Oracle contrast is significant
(41.7\% versus 83.3\%, $p=0.0003$).

Together, the Oracle ladder establishes that perfect spatial information is not
itself sufficient. The dominant gain arises only when the reference can act on
the policy output, while an independent terminal interface resolves a separate
failure dimension. Because these interfaces can also impose an incorrect
reference repeatedly, their benefit creates the need for reliability-aware
authorisation online.

\subsection{Online Reliability Deficit}
\label{sec:online}
Replacing exact bearing and distance with online registration reduces
$\SRret$ to $27/49=55.1\%$. On the 42 episodes with common Outbound success,
online and full Oracle achieve 59.5\% and 85.7\%, respectively, with 16
Oracle-to-online losses and five gains ($p=0.027$). The online system remains
26.2 percentage points below full Oracle on the paired subset. The
online system nevertheless improves substantially over language-only Return:
on 49 common episodes it raises success from 22.4\% to 55.1\% ($p=0.0015$).

The deficit falls almost entirely on arrival. Relative to full Oracle, online
arrival decreases from 90.7\% to 59.2\% (31.5 points), whereas the termination
deficit remains similar (4.7 versus 4.1 points). This does not mean that online
distance is accurate; an incorrect accepted STOP outside the start region is
counted as an arrival failure rather than as arrival without termination.

\begin{figure}[t]
    \centering
    \includegraphics[width=0.96\columnwidth]{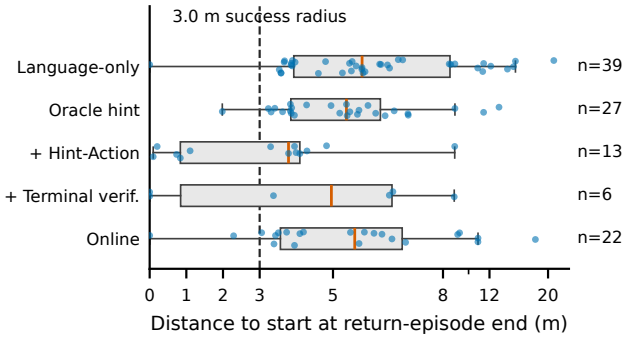}
    \caption{Final distance to $s_0$ for episodes that succeed Outbound but
    fail Return. The dashed line is the 3.0~m radius. Points to its left arrive
    without valid termination; those to its right fail to arrive.}
    \label{fig:failure_distance}
\end{figure}

The final-distance distribution supports the same interpretation
(Fig.~\ref{fig:failure_distance}). Among 22 online Return failures, the median
distance to $s_0$ is 5.59~m, the maximum is 17.91~m, and three runs end beyond
10~m. This resembles Oracle Hint failures (median 5.37~m), where accurate
information is not enforced, rather than the few full-Oracle failures clustered
at or inside the success radius. Under exact information the residual mode is
primarily unstable arrival judgement; online, failure to arrive becomes
dominant again.

Per-step reason codes identify relocalisation confidence as the
dominant restriction: it is below the 0.90 threshold at 1,009 of 1,579 Return
decisions, and at 979 steps (62.0\%) this withholds both the hint and arbiter. Consequently, the hint and action arbiter are unavailable for most of the
online Return trajectory, and the median per-episode override rate falls from
12.6\% under the Oracle condition to 6.9\% online. This is not a narrow threshold-calibration effect: only 30 withheld
steps (3.0\%) lie in $[0.85,0.90)$, while the median relocalisation confidence
over all Return steps is 0.667. Route-state reliability blocks only 0.8\% of all decisions,
although it vetoes 8.5\% of otherwise eligible overrides.For post-hoc evaluation only, we use simulator pose to compute the ground-truth bearing to the target anchor; this information is never provided to the online controller. Of the 130 overrides executed solely from online geometric estimates, 124 (95.4\%) agree with the ground-truth directional class. Among the 12 candidate overrides rejected by the route-state reliability guard, post-hoc evaluation shows that 10 would have been incorrect and only two correct. The guard therefore prevents ten incorrect interventions at the cost of suppressing two correct ones, indicating that it selectively reduces harmful overrides rather than merely reducing their number.

The terminal interface exhibits the complementary failure mode. In 18 of the 22
failed online Return phases, an accepted STOP terminates execution while
post-hoc evaluation places the robot outside the true 3.0~m success region; only four failures are
timeouts. The online deficit is therefore composed primarily
of directional \emph{abstention}, which prevents arrival, together with
over-acceptance by the terminal interface. Improving the fraction of the route
for which geometric evidence can be trusted is more important than further
strengthening interfaces that are already effective under exact information.

Among the 600 online steps on which bearing evidence is authorised, 108 are
skipped near an anchor. Of the remaining 492 decisions, 52.0\% conflict with
the estimated hint, compared with 35.1\% under the Oracle. This does not imply
that the base policy degrades online, because disagreement may also reflect an
incorrect estimate. It does show why reliability cannot be replaced by a rule
that always enforces the online bearing: a locally plausible but wrong anchor
could impose a sustained sequence of incorrect corrections.

\subsection{Alternative Controller: Geometry-Only Route Following}
The Oracle and online experiments above retain NaVILA as the primary
controller and use route memory to support its information, action, and
termination decisions. This raises a natural alternative: given that the online
system already estimates route state, can the VLM be removed entirely and the
executed Outbound route be followed by a deterministic geometric controller?
The experiment does not support this alternative. The geometry-only follower
succeeds on only \mbox{$11/46=23.9\%$} of the 46 valid episodes that complete Outbound.
On the 46 episodes with common Outbound success under geometry-only
and online control, the online system achieves \mbox{$25/46=54.3\%$}, compared
with \mbox{$11/46=23.9\%$} for geometry-only control, a paired advantage of about
30 percentage points (\mbox{$p=0.009$}). Replacing the online hybrid system with this VLM-free deterministic
controller therefore reduces, rather than further improves, Return success.

The geometry-only controller also provides no significant advantage
over Language-only. On the 46 episodes with common Outbound success, their
success rates are \mbox{$23.9\%$} and \mbox{$8/46=17.4\%$}, respectively. Eleven episodes
change from Language-only failure to geometry-only success, whereas eight
change in the opposite direction (\mbox{$p=0.648$}). Taken together, neither the
semantic policy nor simple geometric route following is reliable in isolation;
the value of route memory lies in supporting and constraining the semantic
policy rather than replacing it. Because this alternative configuration also
disables the additional stuck-recovery strategy used by the online system, the
comparison measures the advantage of the complete hybrid system over this
geometric alternative controller and cannot be interpreted as the isolated
causal contribution of the VLM. Stuck recovery is, however, unlikely
to explain the gap: it fired in only 2 of the 27 successful online round trips,
so removing it would leave the online advantage substantially intact.

\section{Discussion and Conclusion}
The experiments expose three separable limits in round-trip VLN. The
standalone-versus-Return comparison localises an additional burden to
continuous execution. Language-only and geometry-only route following
do not differ significantly, whereas the complete online system significantly
outperforms the geometric alternative controller. Because that controller also
has stuck recovery disabled, we report this as an end-to-end comparison rather
than an isolated measurement of the semantic policy's contribution.
The Oracle ladder then shows that information quality is
not the only binding constraint: the present policy conflicts with a correct
reference frequently enough that selective action arbitration produces the
largest gain. Two features of the Oracle ladder deserve emphasis. First, once the
route is acted upon, Return success (86.0\%) exceeds the standalone one-way rate
(40\%): the recorded route removes the directional and
termination ambiguity that limits even a single traversal, so acting on the
recorded route makes Return more reliable than a single cold traversal of the
same path. Second, the gain has two distinct sources that should not be
conflated: arbitration (22.0\% to 70.5\%) is a behavioural gain from acting on
the route, whereas the terminal increment (70.5\% to 86.0\%) partly reflects
forced termination whose distance estimate is thresholded at the evaluation
radius. Return does not reach 100\% because arbitration corrects only about one
decision in six and abstains elsewhere, and because arrival without a supported
STOP still fails. Finally, online evaluation shows that this mechanism depends
on how often the geometric evidence is reliable enough to authorise it. This
sequence matters: evaluating only the final online system would conflate a weak
interface with an effective interface that is usually forced to abstain.

Several limitations qualify the absolute performance. All measurements use one
base policy and one hint template, so the observed conflict rate may not
transfer to models trained to consume structured route information.
The geometry-only baseline represents one deterministic
reverse-polyline follower and also disables stuck recovery. It therefore cannot
establish that every geometric controller requires a VLM, nor can the entire
difference from the online system be attributed to the semantic policy. M50 is
selected for historical Outbound success, and reverse-paired paths retain a
non-zero endpoint offset; paired contrasts are more robust than absolute rates.
The route-state prior is maintained by dead reckoning over commanded
action deltas rather than measured or leg odometry, which the online system does
not consume. The three reliability components are combined by a conservative
joint (AND) gate rather than authorising separate operations, so the present
results do not isolate the value of per-operation specialisation against a
single shared confidence; that separation, and a no-gate control, are left to
future work. Existing logs already show that the joint gate reduces wrong
interventions (Sec.~\mbox{\ref{sec:online}}).
The sample contains approximately
40 paired Return episodes per contrast, leaving the smaller terminal increment
underpowered. Finally, evaluation is limited to simulation and a single robot
platform.

We have presented a reliability-aware sparse memory of the route executed during
Outbound and a diagnostic framework that separates directional information,
behavioural consistency, and termination. The same routes run as standalone one-way
episodes succeed 40\%, whereas language-only Return and
deterministic geometric replay achieve conditional success rates of only 22.0\%
and 23.9\%, showing that continuous Return is harder than one-way execution and
that neither the semantic policy nor geometric replay is sufficient in
isolation. Under exact information, language
alone provides only a modest, non-significant improvement, whereas selective
action arbitration and terminal verification raise Return success to 86.0\%.
Online matching retains a significant improvement over the baseline but loses
26.2 paired points to the Oracle, principally because bearing uncertainty
withholds intervention. Future work should replace single-frame local matching
with temporal, multimodal route belief and active disambiguation, and test the
result across policies and on a physical mobile robot.

\bibliographystyle{IEEEtran}
\bibliography{references}
\end{document}